\documentclass[10pt,twocolumn,letterpaper]{article}

\usepackage{booktabs} % For formal tables
\def\vector#1{\mbox{\boldmath $#1$}}
\usepackage{multirow}

\usepackage{times}
\usepackage{epsfig}
\usepackage{graphicx}
\usepackage{amsmath}
\usepackage{amssymb}

\usepackage{color}

\usepackage{array} % For table resize

\usepackage[pagebackref=true,breaklinks=true,letterpaper=true,colorlinks,bookmarks=false]{hyperref}

\usepackage[font={small,it}]{caption}

\begin{document}

%%%%%%%%% TITLE
\title{Conditional Video Generation Using Action-Appearance Captions}

\author{{S. Yamamoto}$^1$
\and
{A. Tejero-de-Pablos}$^1$
\and
{Y. Ushiku}$^1$
\and
{T. Harada}$^{1,2}$
}
\date{
$^1$The University of Tokyo\hspace{5mm}
$^2$RIKEN%\\[2ex]%
}
%\date{}
\maketitle
%\thispagestyle{empty}

%%%%%%%%% ABSTRACT
\begin{abstract}
\textit{The field of automatic video generation has received a boost thanks to the recent Generative Adversarial Networks (GANs).
However, most existing methods cannot control the contents of the generated video using a text caption, losing their usefulness to a large extent. This particularly affects human videos due to their great variety of actions and appearances.
This paper presents Conditional Flow and Texture GAN (CFT-GAN), a GAN-based video generation method from action-appearance captions. We propose a novel way of generating video by encoding a caption (e.g., ``\textnormal{a man in blue jeans is playing golf}'') in a two-stage generation pipeline. Our CFT-GAN uses such caption to generate an optical flow (action) and a texture (appearance) for each frame. As a result, the output video reflects the content specified in the caption in a plausible way. Moreover, to train our method, we constructed a new dataset for human video generation with captions. We evaluated the proposed method qualitatively and quantitatively via an ablation study and a user study. The results demonstrate that CFT-GAN is able to successfully generate videos containing the action and appearances indicated in the captions.}
\end{abstract}

%%%%%%%%% BODY TEXT
\vspace{-5mm}
\section{Introduction}

The field of multimedia content creation has experienced a remarkable evolution since the application of Generative Adversarial Networks (GANs) to image and video generation. Radford et al. \cite{radford2015unsupervised} succeeded in generating realistic images using Deep Convolutional Generative Adversarial Networks (DCGANs). As an improvement, Wang et al. \cite{wang2016generative} proposed a GAN that takes into account structure and style. Using a two-stage architecture, they achieved more plausible images.
These methods are able to generate images randomly but do not allow to specify the image content. To approach this problem, Zhang et al. \cite{zhang2016stackgan} proposed StackGAN to generate photo-realistic images from a description sentence or {\it caption} (e.g., ``{\it a man standing in a park}'').

\begin{figure}[t]
\begin{center}
  	\includegraphics[width=\hsize]{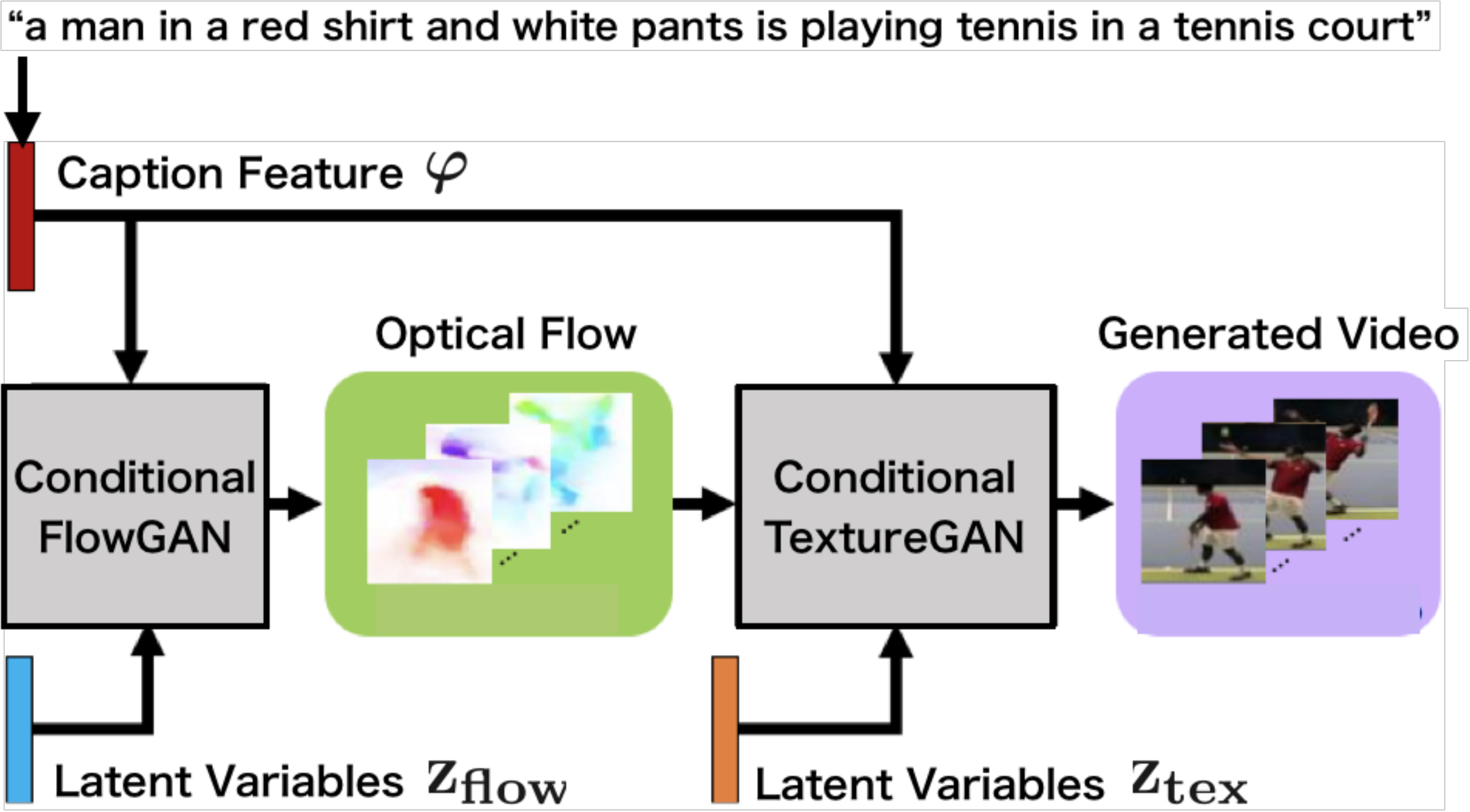}
\end{center}
   \vspace{-2mm}
   \caption{Overview of the proposed method, CFT-GAN. We employ a two-stage generative architecture: Conditional FlowGAN generates optical flow from latent variables $\vector{z}_{\rm{flow}}$ and features $\varphi$ extracted from the input caption; Conditional TextureGAN generates video from latent variables $\vector{z}_{\rm{tex}}$, features $\varphi$ extracted from the input caption and the generated optical flow.}
   \label{fig:overview}
   \vspace{-5mm}
\end{figure}

In addition to images, automatic generation of video content using GANs has also been studied. Video generation is a more difficult task, since the content between frames has to be consistent, and the motion of objects has to be plausible.
This is particularly challenging in the case of human video, due to the complexity of actions and appearances.
Vondrick et al. \cite{vondrick2016generating} proposed a scene-consistent video generation method (VGAN). The videos generated by VGAN have a consistent background, but the motion of humans is usually distorted and not realistic. To improve this, Ohnishi et al. \cite{ohnishi2018ftgan} proposed a method for generation of realistic video by employing optical flow in a two-stage pipeline to improve the plausibility of actions.
To control the content of the generated video, many works provide an image (the first frame) as a condition \cite{hao2018controllable, yang2018pose}, but few of them use captions \cite{pan2017create, marwah2017attentive}. Providing captions requires almost no effort, and results are potentially more creative than using an input image.
Moreover, previous methods show little variety of human actions and appearances. In order to overcome this problem, we explored the way captions are encoded into the generation pipeline, aiming for a video generator that can be controlled to reflect a variety of actions and appearances.

%However, none of these methods allow to control the content of the generated video, and therefore they lose their usefulness to a large extent.
%While GAN-based automatic image generation from captions has been studied, to the best of our knowledge, GAN-based video generation methods from captions have not been proposed yet.

In this paper, we present a novel video generation method from action-appearance captions: Conditional Flow and Texture GAN (CFT-GAN). An action-appearance caption is a sentence describing a subject, the action performed, and the background (e.g., {\it a man in blue jeans is playing golf in a field}).
We propose a way of encoding caption features as a condition to generate both the optical flow and the final video.
In order for our videos to show plausible actions, we first generate an optical flow to represent the motion in the scene, as in \cite{ohnishi2018ftgan}.
Figure \ref{fig:overview} shows system overview of CFT-GAN. Our method consists of two components: Conditional FlowGAN generates the motion of the scene using an action-appearance caption as a condition; Conditional TextureGAN generates the output video using the same caption and the motion generated by Conditional FlowGAN as a condition. To the best of our knowledge, this is the first GAN-based method for video generation from action-appearance captions.

We also constructed a new dataset\footnote{The dataset will be released after publication.} for video generation from action-appearance captions, and used it to evaluate our method. To the best of our knowledge, there is no such dataset available, so we captioned a human action video dataset for our generation purposes.

The contributions of this paper are as follows.
\begin{itemize}
\item We propose a novel method for automatic video generation from captions, CFT-GAN. Our way of encoding caption features into a two-stage architecture allows us to control the action and appearance of the generated video.
\item We constructed a new video dataset with action-appearance captions, and used it to train our method. We also explain how to properly train the complex architecture of CFT-GAN.
\item We provide an evaluation of different caption encodings via an ablation study, and a verification via a user study.
\end{itemize}

\section{Related work}
\label{sec:related}
%%%画像生成の話
The field of automatic image and video generation has experienced a boost due to the emergence of Generative Adversarial Networks (GANs) \cite{goodfellow2014generative,radford2015unsupervised}.
%Previous content generation methods were based on Variational Auto-Encoders (VAEs) \cite{kingma2013auto}. However, the VAE loss function causes generated images to resemble the average of the training data and to look blurry. In comparison, recent GAN-based methods perform better than VAE for image and video generation tasks.
Unlike previous methods (i.e., Variational Auto-Encoders), GANs allow generating frames not contained in the original dataset.
%%%Image generation
In image generation, Pix2Pix \cite{isola2016image} proposed an architecture based on a U-net network \cite{ronneberger2015u} to convert an input image to a target image that shares the same edges. The bypass between the upper-layers and lower-layers of U-net allows the output image to reflect spatial information from the input (e.g, edges).
In order to improve the realism of the generated images, Style and Structure GAN (${\rm S^{2}}$GAN) \cite{wang2016generative} relies on a two-stage generation method to preserve the structure of the objects. First, Structure-GAN generates the underlying 3D model; then, the 3D model is input to Style-GAN, which generates the output 2D image. %In ${\rm S^{2}}$GAN, it is important to consider \textcolor{red}{WHAT'S THIS: fundamental information of target domains} for generation.
%%%Conditional image generation
Since the aforementioned works do not provide a way of controlling the generated image, methods to impose a condition in the output content have also been studied \cite{zhang2016stackgan, reed2016generative, odena2017conditional, kaneko2018generative, xu2018attngan}. StackGAN \cite{zhang2016stackgan} is able to generate realistic images from captions. It extracts text features from captions, and uses them as a condition for a two-stage generator. The first-stage generator generates a low resolution image from a set of latent variables and the caption features. Then, the second-stage generator generates a high resolution image using the same latent variables and caption features, and the low resolution image generated by the first-stage generator.

%%%動画像生成の話(VGAN)
The task of automatic video generation has been also approached using GANs. However, video generation is more challenging, since it requires consistency between frames and motion should be plausible. This is particularly challenging in the case of human motion generation.
Video GAN (VGAN) \cite{vondrick2016generating} achieves scene-consistent videos by generating the foreground and background separately. This method consists of 3D convolutions that learn motion information and appearance information simultaneously. However, capturing both motion and appearance using single-stream 3D convolutional networks causes generated videos to have problems with either their visual appearance or motion.
%For example, some generated videos have consistent scenes but no motion, while in others the movements are not plausible.
%%%動画像生成の話(FTGAN)
Recent methods \cite{ohnishi2018ftgan, tulyakov2018mocogan} explore the fact that videos consist of motion and appearance. In \cite{ohnishi2018ftgan}, a hierarchical video generation system is proposed: Flow and Texture GAN (FTGAN). FTGAN consists of two components: FlowGAN generates the motion of the video, which is used by TextureGAN to generate videos. This method is able to successfully generate realistic video that contains plausible motion and consistent scenes.

%%% Updated state of the art
Although in \cite{vondrick2016generating, ohnishi2018ftgan, tulyakov2018mocogan} there is no way to control the content of the videos, some other methods have attempted to condition video generation. In \cite{hao2018controllable, yang2018pose, zhao2018learning, cai2018deep, he2018probabilistic}, the video is generated by providing the first frame of the sequence as the reference. These works also fall into the category of video prediction, since they require providing the initial state of the scene.
%As in \cite{ohnishi2018ftgan, tulyakov2018mocogan}, \cite{yang2018pose, zhao2018learning} use a two-stage generation pipeline to preserve coherency between human motion and scene appearance.
While providing an image constricts the degrees of freedom of the generated video, this may be undesirable or impractical in certain creative applications.

We believe that captions can be leveraged as an alternative, more practical way of controlling the content of the generated video. To the best of our knowledge, there is comparatively few works on automatic video generation from captions \cite{pan2017create, marwah2017attentive, saito2017temporal}. However, whereas \cite{pan2017create} does not tackle the challenge of generating human actions, \cite{marwah2017attentive} only handles simple movements such as ``walking left'' or ``running right'' in a black-and-white scene. TGAN \cite{saito2017temporal} handles more complex human actions, using an action-class word (i.e., \textit{golf}) to condition the generated video. These methods' architecture is single-stage.
%Also, neither of them utilizes a two-stage pipeline for appearance-motion coherence.}

While image generation methods became able to reflect the content of an input caption, video generation has not achieved that level of control yet. We believe that, by leveraging the two-stage architecture for texture and motion, it is possible to condition, not only the type of action, but also human appearance and background. In this paper, we propose a method for automatic video generation based on the encoding caption features into the video's motion and appearance. To the best of our knowledge, this is the first method capable of generating a video that reflects the human action and appearance specified in an input caption.

\begin{figure*}[t]
\begin{center}
  	\includegraphics[scale=0.25]{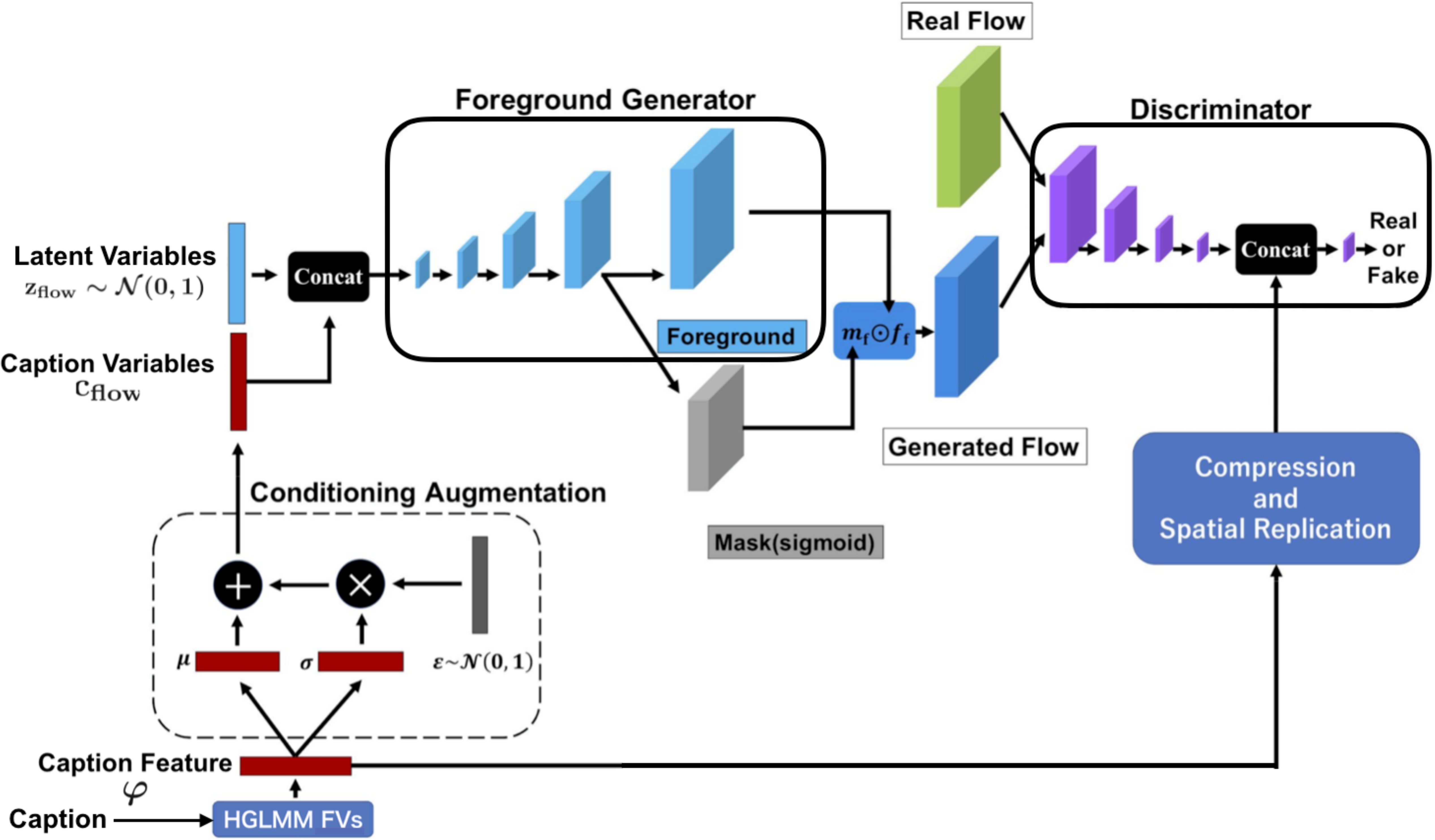}
\end{center}
   \vspace{-1mm}
   \caption{Architecture of Conditional FlowGAN (training): Given the latent variables $\vector{z}_{\rm{flow}}$ sampled from Gaussian distributions and the caption as input, the generator learns to generate the optical flow that represents the action specified in the caption. Real and generated optical flows have a resolution of $64 \times 64$ pixels and a duration of 32 frames.}
   \label{fig:cflowgan}
   \vspace{-5mm}
\end{figure*}

\section{Preliminaries}
\subsection{Generative adversarial networks}
Generative Adversarial Networks (GANs) \cite{goodfellow2014generative} consist of two networks: a generator ($G$) and a discriminator ($D$). $G$ attempts to generate data that looks similar to the given dataset. The input for $G$ is a latent variable \vector{z}, which is randomly sampled from a probability distribution $p_{\vector{z}}$ (e.g., a Gaussian distribution).

$D$ attempts to distinguish between data from the dataset (real data) and data generated from $G$ (generated data). During training, a GAN simultaneously updates these two networks according to the following objective function $V$:
\begin{equation}
\begin{split}
	\min_{G} \max_{D} V(G,D) &= \mathbb{E}_{\vector{x} \sim dataset} [\log(D(\vector{x}))] \\
	&+ \mathbb{E}_{\vector{z} \sim p_{\vector{z}}} [\log (1-D(G(\vector{z})))] 
\end{split}
\end{equation}
where $\vector{x}$ is the data from the dataset (real data).

\subsection{Single-stage GAN for video generation}
Generative Adversarial Network for Video (VGAN) \cite{vondrick2016generating} is a network for video generation based on the concept of GANs; it consists of a generator and a discriminator. The VGAN generator comprises a mask architecture to generate separately a static background and a moving foreground from latent variables $z$:
\begin{equation}
	G(\vector{z}) = m(\vector{z}) \odot f(\vector{z}) + (1 - m(\vector{z})) \odot b(\vector{z})
\end{equation}
where $\odot$ represents the element-wise multiplication, and $m(\vector{z})$ is a spatiotemporal matrix with values in the range of 0 to 1. For each pixel $(\vector{x},\vector{y},\vector{t})$, the mask selects whether the foreground $f(\vector{z}$) or the background $b(\vector{z})$ appears in the video. To generate a consistent background, $b(\vector{z})$ produces a spatial static image replicated over time. During training, in order to emphasize the background image, L1 regularization $\lambda \|m(\vector{z})\|_{1}$ for $\lambda = 0.1$ is added to the GAN objective function. 

\subsection{Two-stage GAN for video generation}
Hierarchical video generation networks (e.g., FTGAN) \cite{ohnishi2018ftgan} are based on the fact that videos consist of two elements: motion and appearance. Likewise, FTGAN consists of two components: FlowGAN and TextureGAN. FlowGAN generates motion in the form of optical flow from latent variables. Then, TextureGAN generates videos from latent variables and the optical flow generated by FlowGAN.

\subsubsection{FlowGAN}
FlowGAN generates optical flow from latent variables $\vector{z}_{\rm{flow}}$. The architecture of FlowGAN is based on VGAN \cite{vondrick2016generating}. VGAN is able to generate scene-consistent videos by generating the foreground and background separately. However, considering that the value of the optical flow should be zero for a static background, the FlowGAN generator does not need to learn a background stream. Instead, they make $b(\vector{z})$ a zero matrix, which is equivalent to using only the foreground stream of VGAN. Therefore, in FlowGAN, optical flow $G_{\rm{flow}}$ is generated as follows:
\begin{equation}
G_{\rm{flow}}\left(\vector{z}_{\rm{flow}} \right) = m\left(\vector{z}_{\rm{flow}} \right) \odot f\left( \vector{z}_{\rm{flow}} \right)
\end{equation}

\subsubsection{TextureGAN}
TextureGAN takes the optical flow generated by FlowGAN and latent variables $\vector{z}_{\rm{tex}}$ as input and generates the output video. The architecture of the generator is based on Pix2Pix \cite{isola2016image} and VGAN, which generates foreground and background separately. The foreground generator is based in the U-net architecture \cite{ronneberger2015u}, as in Pix2Pix. The bypasses between upper and lower layers in U-net allow reflecting the spatial information of the input into the output. Thus, the sharp edges of the input optical flow are reflected as the shapes of the moving objects/humans in the foreground of generated video. In TextureGAN, video $G_{\rm{tex}}$ is generated as follows:
\begin{equation}
\begin{split}
	G_{\rm{tex}}\left(\vector{z}_{\rm{tex}},\vector{f}\right) &= m\left(\vector{z}_{\rm{tex}},\vector{c}\right) \odot f\left( \vector{z}_{\rm{tex}},\vector{f} \right) \\
	&\quad+ \left(1- m\left(\vector{z}_{\rm{tex}},\vector{f}\right) \right) \odot b\left(\vector{z}_{\rm{tex}}\right)
\end{split}
\end{equation}
where $\vector{f}$ is the input optical flow. Apart from using the optical flow generated by FlowGAN in the foreground generator, the \textit{ground truth} optical flow is used to train the discriminator.

\subsection{Conditional GAN for image generation}

One of the most prominent methods for including a caption as a condition to generate images is StackGAN \cite{zhang2016stackgan}.
Given an input caption (e.g., ``{\it a gray bird with white on its chest and a very short beak}''), StackGAN extracts a feature embedding from the text and uses it, along with the latent variables, for generating the image (a two-stage generator, see Section \ref{sec:related}).
However, the limited number of training pairs of images and captions often results in sparsity in the text conditioning manifold. Such sparsity makes training a GAN difficult. To solve this problem, StackGAN introduces a conditioning augmentation technique. Instead of directly using the raw caption embedding $\varphi$ as caption features, they use latent variables randomly sampled from an independent Gaussian distribution $\mathcal{N}(\mu(\varphi),\sigma(\varphi))$, where the mean $\mu(\varphi)$ and diagonal covariance matrix $\sigma(\varphi)$ are functions of the caption embedding $\varphi$. This conditioning augmentation technique smooths the distribution of caption features and makes StackGAN relatively easy to train.

\begin{figure*}[t]
\begin{center}
  	\includegraphics[scale=0.45]{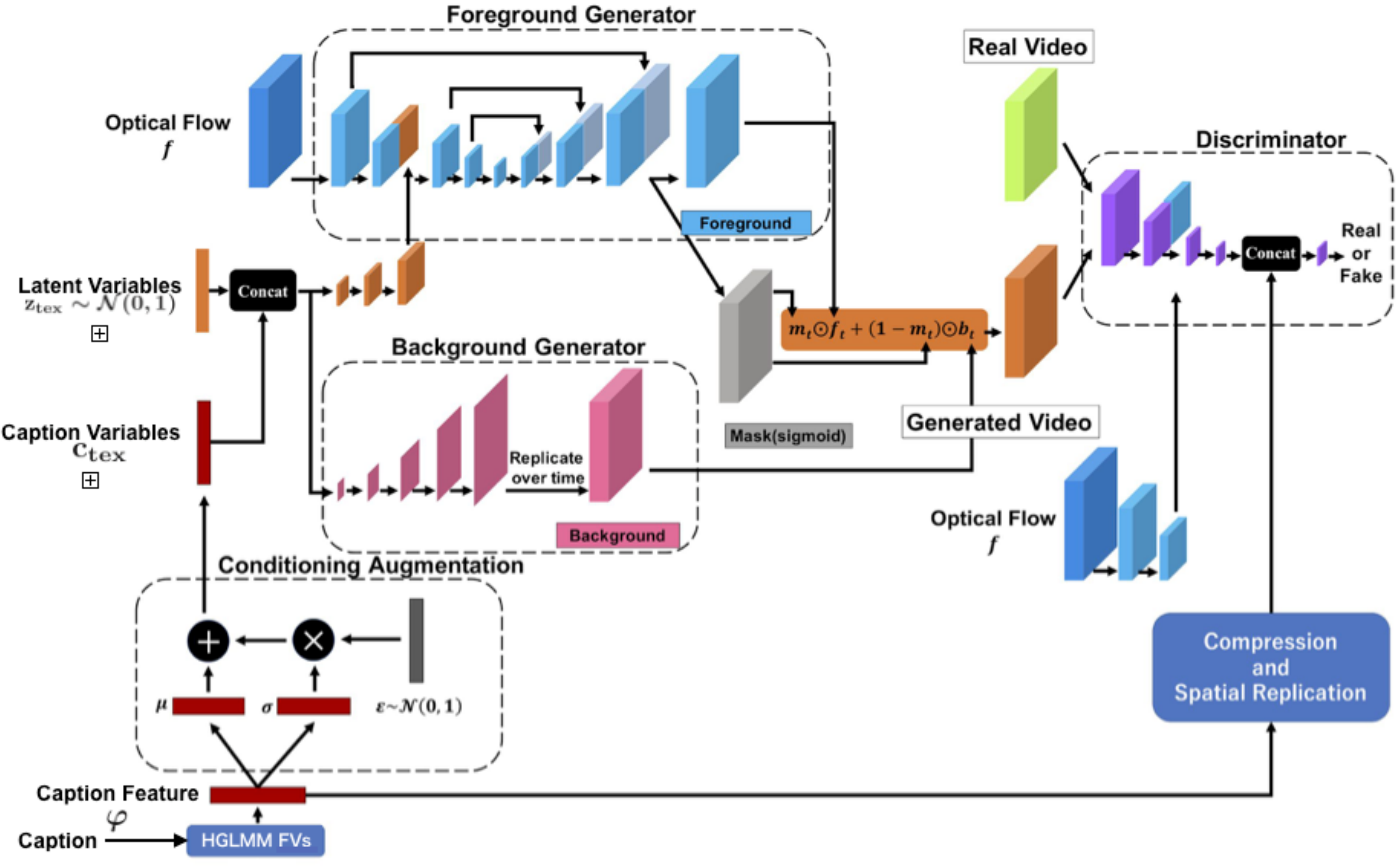}
\end{center}
   \vspace{-1mm}
   \caption{Architecture of Conditional TextureGAN (training): Given the optical flow $f$, the latent variables $z_{tex}$ sampled from Gaussian distributions, and the caption as input, the generator learns to generate the video that represents the action and appearance specified in the caption. The input video, the input optical flow, and the output video have a resolution of $64 \times 64$ pixels and a duration of 32 frames.}
   \label{fig:ctexgan}
   \vspace{-5mm}
\end{figure*}

\section{Video generation from action-appearance captions}

We propose Conditional Flow and Texture GAN (CFT-GAN), a novel method for video generation. As in \cite{zhang2016stackgan}, our method uses features extracted from an input caption as a condition for the generated content. To ensure the presence of motion and visual details in the generated video, we employ action-appearance captions, that is, captions that express an action, the appearance of the subject, and the background (e.g., ``{\it a lady in a black dress doing sit ups in the gym}'').
In order for the video to reflect the action in a plausible way, CFT-GAN separates the video generation hierarchically in two stages, as in \cite{ohnishi2018ftgan}: Conditional FlowGAN generates optical flow motion based on the input caption; Conditional TextureGAN generates the output video using both the input caption and the generated optical flow as a condition.

\subsection{Conditional FlowGAN}
Conditional FlowGAN generates optical flow from caption features $\varphi$, and latent variables $\vector{z}_{\rm{flow}}$. Figure \ref{fig:cflowgan} shows the architecture of Conditional FlowGAN.

First, to extract features $\varphi$ from the input caption, we use Fisher Vectors (FVs) \cite{perronnin2007fisher} based on a hybrid Gaussian-Laplacian mixture model (HGLMM) \cite{klein2015associating}. Then, as in \cite{zhang2016stackgan}, we extract caption variables $\vector{c}_{\rm{flow}}$ using conditioning augmentation. That is, we calculate the mean $\mu(\varphi)$ and diagonal covariance matrix $\sigma(\varphi)$ from our caption features $\varphi$, and randomly sample latent variables $\vector{c}_{\rm{flow}}$ from an independent Gaussian distribution $\mathcal{N}(\mu(\varphi),\sigma(\varphi))$.

Then, our latent variables $\vector{z}_{\rm{flow}}$ are sampled from an independent Gaussian distribution, where $\mu=0$ and $\sigma=1$, i.e. $\mathcal{N}(0,1)$. After that, we concatenate our caption variables $\vector{c}_{\rm{flow}}$ and our latent variables $\vector{z}_{\rm{flow}}$, and generate the foreground of the optical flow $\vector{f}_{\rm{f}}$ and the mask of the optical flow $\vector{m}_{\rm{f}}$. As in \cite{vondrick2016generating, ohnishi2018ftgan}, the mask $\vector{m}_{\rm{f}}$ is a spatiotemporal matrix with each value ranging from 0 to 1; it selects either the foreground $\vector{f}_{\rm{f}}$ or the background for each pixel $(\vector{x},\vector{y},\vector{t})$. According to \cite{ohnishi2018ftgan}, the background of optical flow should be zero if the camera is fixed, so Conditional FlowGAN does not require to learn a background generator. Instead, we use a zero matrix as background.
We merge the foreground $\vector{f}_{\rm{f}}$ and the background (zero matrix) based on the mask $\vector{m}_{\rm{f}}$ as follows:
\iffalse
\begin{eqnarray}
	G_{\rm{flow}} (\vector{z}_{\rm{flow}},\vector{c}_{\rm{flow}}) = \vector{m}_{\rm{f}}(\vector{z}_{\rm{flow}},\vector{c}_{\rm{flow}})  \odot \vector{f}_{\rm{f}}(\vector{z}_{\rm{flow}},\vector{c}_{\rm{flow}}) 
\end{eqnarray}
\fi
\begin{align}
	G_{\rm{flow}} {}&(\vector{z}_{\rm{flow}},\vector{c}_{\rm{flow}}) =  \notag {}\\ &\vector{m}_{\rm{f}}(\vector{z}_{\rm{flow}},\vector{c}_{\rm{flow}})  \odot \vector{f}_{\rm{f}}(\vector{z}_{\rm{flow}},\vector{c}_{\rm{flow}}) 
\end{align}

Finally, for the discriminator, we compressed the caption features $\varphi$ using a fully-connected layer and then replicated them spatially, and concatenated the result to one of the middle layers of the discriminator, as in \cite{zhang2016stackgan}. By doing this, the discriminator can judge not only whether the input optical flow is real or generated, but also whether the pair of input optical flow and caption are plausible.

\subsection{Conditional TextureGAN}
As shown in Figure \ref{fig:ctexgan}, Conditional TextureGAN generates video from caption features, latent variables $\vector{z}_{\rm{tex}}$, and the optical flow $\vector{f}$ generated by Conditional FlowGAN.

First, we extract caption features $\varphi$ via HGLMM FVs as in Conditional FlowGAN, and then extract caption variables $\vector{c}_{\rm{tex}}$ using conditioning augmentation. After that, we calculate the spatiotemporal matrix from caption variables $\vector{c}_{\rm{tex}}$ via two up-sampling blocks. 

We input the optical flow generated by FlowGAN to the foreground generator, which has a U-net structure \cite{ronneberger2015u}. The U-net structure contains a bypass between upper and lower layers that allows reflecting the spatial information of the input (i.e., the sharp edges of the optical flow) into the output (i.e., the foreground and mask).

We input the spatiotemporal matrix calculated from caption variables $\vector{c}_{\rm{tex}}$ to the foreground generator at one of the middle layers to generate the foreground of the video $\vector{f}_{\rm{t}}$ and the mask of the video $\vector{m}_{\rm{t}}$. The mask $\vector{m}_{\rm{t}}$ is a spatiotemporal matrix with values in the range of 0 to 1; it selects either the foreground $\vector{f}_{\rm{t}}$ or the background $\vector{b}_{\rm{t}}$ for each pixel (x, y, t).
Simultaneously, the background generator generates the background $\vector{b}_{\rm{t}}$ from the concatenation of $\vector{c}_{\rm{tex}}$ and $\vector{z}_{\rm{tex}}$.
Finally, we merge the foreground $\vector{f}_{\rm{t}}$ and the background $\vector{b}_{\rm{t}}$ according to the mask $\vector{m}_{\rm{t}}$ as follows:
\iffalse
\begin{equation}
\begin{split}
	G_{\rm{tex}} (\vector{z}_{\rm{tex}},\vector{c}_{\rm{tex}},\vector{f}) = \vector{m}_{\rm{t}}(\vector{z}_{\rm{tex}},\vector{c}_{\rm{tex}},\vector{f})  \odot \vector{f}_{\rm{t}}(\vector{z}_{\rm{tex}},\vector{c}_{\rm{tex}},\vector{f}) \\
    	+ \left(1- \vector{m}_{\rm{ t}}\left(\vector{z}_{\rm{tex}},\vector{c}_{\rm{tex}},\vector{f}\right) \right) \odot \vector{b}_{\rm{t}}\left(\vector{z}_{\rm{tex}},\vector{c}_{\rm{tex}}\right)
        \end{split}
\end{equation}
\fi
\begin{align}
	G_{\rm{tex}} {}&(\vector{z}_{\rm{tex}},\vector{c}_{\rm{tex}},\vector{f}) =  \notag {}\\ &\vector{m}_{\rm{t}}(\vector{z}_{\rm{tex}},\vector{c}_{\rm{tex}},\vector{f})  \odot \vector{f}_{\rm{t}}(\vector{z}_{\rm{tex}},\vector{c}_{\rm{tex}},\vector{f}) \notag {}\\
    	&+ \left(1- \vector{m}_{\rm{ t}}\left(\vector{z}_{\rm{tex}},\vector{c}_{\rm{tex}},\vector{f}\right) \right) \odot \vector{b}_{\rm{t}}\left(\vector{z}_{\rm{tex}},\vector{c}_{\rm{tex}}\right)
\end{align}

Calculating the foreground and the background of the videos separately allows generating scene-consistent videos. Also, using motion information (optical flow) increases the plausibility of the actions contained in the generated video.

For the discriminator, we concatenated the spatiotemporal matrix extracted from the optical flow to the second layer of the discriminator. By doing this, the discriminator can judge not only whether the input video is real or generated, but also whether the pair of input video and optical flow are plausible.
Also, as in Conditional FlowGAN, we compressed the caption features $\varphi$ using a fully-connected layer and then replicated them spatially. Then, we concatenated them to one of the middle layers of the discriminator.

In this way, including the caption features in both Conditional FlowGAN and Conditional TextureGAN allows us to control the motion and the appearance of the video respectively.

\subsection{Implementation details}

To compute the HGLMM-based FVs, we trained an HGLMM with 30 centers using 300-dimensional word vectors \cite{mikolov2013distributed} to extract text descriptors of the caption. Next, we compute the FVs of the descriptors using the learned HGLMM, and then apply principal components analysis (PCA) to reduce their size from 18000 to 256 dimensions. The size of our caption variables $\vector{c}_{\rm{flow}}$ and $\vector{c}_{\rm{tex}}$ is 128 dimensions, and the latent variables $\vector{z}_{\rm{tex}}$ and $\vector{z}_{\rm{flow}}$ are sampled from Gaussian distributions with 100 dimensions.

Conditional FlowGAN and Conditional TextureGAN contain up-sampling blocks and down-sampling blocks. Up-sampling blocks consist of the nearest-neighbor up-sampling  followed by $4 \times 4$ stride 2 convolutions. Conditional FlowGAN has 4 up-sampling blocks in the foreground generator. Conditional TextureGAN has 4 up-sampling blocks in the foreground generator, background generator, and has 2 up-sampling blocks for concatenating $\vector{c}_{\rm{tex}}$ and $\vector{z}_{\rm{tex}}$ to the foreground generator.
Batch normalization \cite{ioffe2015batch} and Rectified linear unit (ReLU) activation are applied after every up-sampling convolution, except for the last layer. The down-sampling blocks consist of $4 \times 4$ stride 2 convolutions, and we apply batch normalization \cite{ioffe2015batch} and LeakyReLU \cite{xu2015empirical} to all layers but only apply batch normalization to the first layer. Conditional FlowGAN has 4 down-sampling blocks in the discriminator. Conditional TextureGAN has 4 down-sampling blocks in both the foreground generator and the discriminator, and 2 down-sampling blocks for extracting the spatiotemporal matrix from the input optical flow in the discriminator.

When training the networks, we use the Adam \cite{kingma2014adam} optimizer with an initial learning rate $\alpha = 0.0002$ and momentum parameter $\beta_{1} = 0.5$. The learning rate is decayed to $1/2$ from its previous value every 10,000 iterations during the training. We set a batch size of $32$.

Although it is desirable to train Conditional FlowGAN and Conditional TextureGAN simultaneously, Conditional TextureGAN cannot be trained unless Conditional FlowGAN has been trained to some extent. For this, we use real optical flow calculated from real videos. Thus, at the beginning of the training, Conditional TextureGAN is updated mainly based on the loss obtained using real optical flow. Then, the network is updated gradually based on the loss obtained using the optical flow generated by Conditional FlowGAN. Loss functions are as follows:
\begin{align}
	{}&L_{D_{\rm{flow}}} = \log D_{\rm{flow}}(\vector{f}) \notag {}\\
	&+\log (1-D_{\rm{flow}}(G_{\rm{flow}}(\vector{z}_{\rm{flow}},\vector{c}_{\rm{flow}}))) %D_flow
\\
\notag \\
    {}&L_{G_{\rm{flow}}} = \log (1-D_{\rm{flow}}(G_{\rm{flow}}(\vector{z}_{\rm{flow}},\vector{c}_{\rm{flow}}))) \notag {}\\ 
    &+\frac{\rm{k}}{\rm{K}} \log (1-D_{\rm{tex}}(G_{\rm{tex}}(\vector{z}_{\rm{tex}},\vector{c}_{\rm{tex}},G_{\rm{flow}}(\vector{z}_{\rm{flow}},\vector{c}_{\rm{flow}})))) %G_flow
\\
    {}&L_{D_{\rm{tex}}} = \log D_{\rm{tex}}(\vector{x}) \notag {}\\
    &+(1 - \frac{\rm{k}}{\rm{K}}) \log (1-D_{\rm{tex}}(G_{\rm{tex}}(\vector{z}_{\rm{tex}},\vector{c}_{\rm{tex}},\vector{f}))) \notag {}\\
	&+\frac{\rm{k}}{\rm{K}} \log (1-D_{\rm{tex}}(G_{\rm{tex}}(\vector{z}_{\rm{tex}},\vector{c}_{\rm{tex}},G_{\rm{flow}}(\vector{z}_{\rm{flow}},\vector{c}_{\rm{flow}}))))
\\
    {}&L_{G_{\rm{tex}}} = (1 - \frac{\rm{k}}{\rm{K}}) \log (1-D_{\rm{tex}}(G_{\rm{tex}}(\vector{z}_{\rm{tex}},\vector{c}_{\rm{tex}},\vector{f}))) \notag {}\\
	&+\frac{\rm{k}}{\rm{K}} \log (1-D_{\rm{tex}}(G_{\rm{tex}}(\vector{z}_{\rm{tex}},\vector{c}_{\rm{tex}},G_{\rm{flow}}(\vector{z}_{\rm{flow}},\vector{c}_{\rm{flow}}))))
\end{align}
where $\rm{k}$ is the number of the current iteration, and $\rm{K}$ is the total number of iterations. We train CFT-GAN for $\rm{K}=60000$ iterations.

\begin{figure}[t]
\begin{center}
  	\includegraphics[width=\hsize]{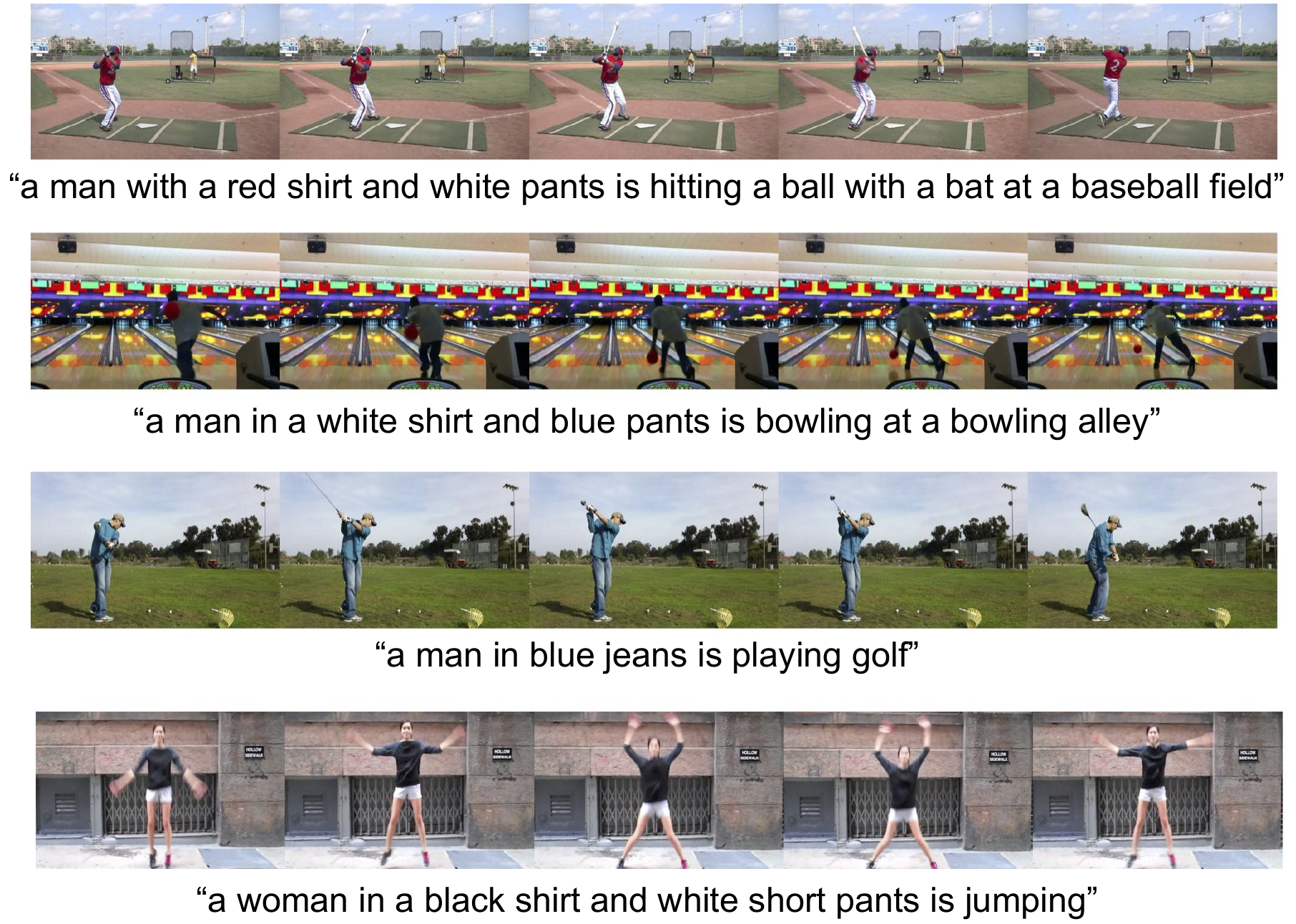}
\end{center}
   \vspace{-3mm}
   \caption{Sample instances of our self-constructed dataset for video generation from action-appearance captions. We added descriptive captions to the Penn Action dataset \cite{zhang2013actemes} via Amazon Mechanical Turk \cite{amt}.}
   \label{fig:dataset}
   \vspace{-5mm}
\end{figure}

\section{Dataset and settings}
We evaluated our method for the task of human video generation due to the variety of actions and appearances featured. 
In order to train our method for this task, a dataset containing videos and their corresponding descriptive captions are necessary. However, to the best of our knowledge, such video dataset does not exist at the moment. Thus, in this study, we constructed a new video dataset for video generation from action-appearance captions by using an existing video dataset and adding captions that describe its content. We used the Penn Action dataset \cite{zhang2013actemes}, which contains 2326 videos of 15 different classes, amounting to a total of 163841 frames. This dataset also contains the position of the body joints of the human shown in each frame; for our dataset, we used these positions to crop the area containing only the human. Then, we used Amazon Mechanical Turk (AMT \cite{amt}) to obtain one descriptive action-appearance caption per video in the dataset. An action-appearance caption contains an action (e.g., ``\textit{is jumping}''), the appearance of the person doing the action (e.g., ``\textit{a man in blue jeans}''), and in some cases the background (e.g., ``\textit{at a baseball field}''). Figure \ref{fig:dataset} shows some sample instances of our dataset.

We used Epic flow \cite{revaud:hal-01142656} as the ground truth optical flow to train our networks, as in \cite{ohnishi2018ftgan}. We resized all frames and optical flow images to a resolution of $76\times76$ pixels, and augmented them by cropping them into $64\times64$ resolution images and randomly applying horizontal flips during training. In addition, since the length of the videos generated by our method and the baseline is 32 frames, we randomly cut 32 frames of each video for training.

\begin{figure}[t]
\begin{center}
  	\includegraphics[width=\hsize]{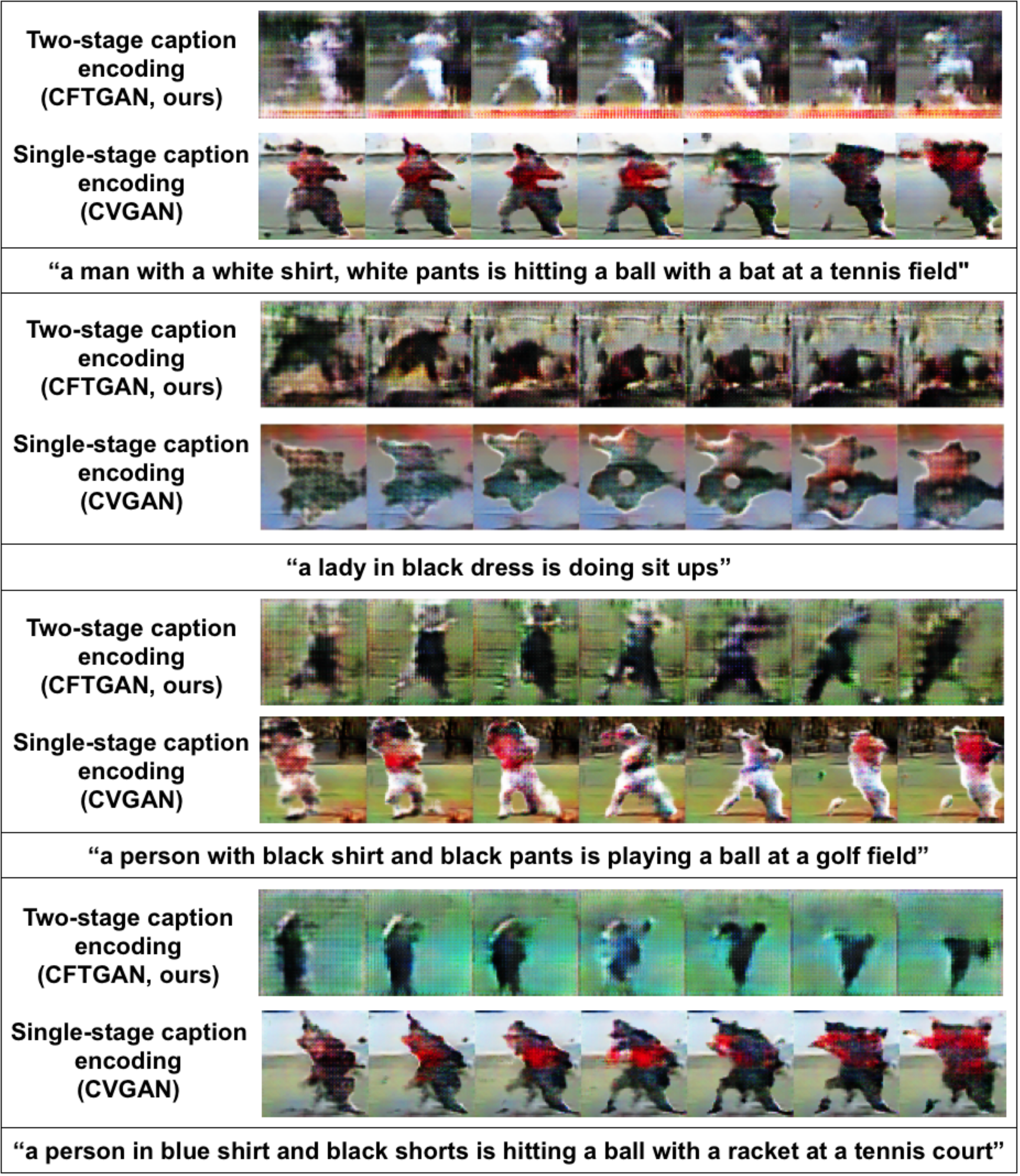}
\end{center}
   %\vspace{4mm}
   \caption{Examples of videos generated by our method (CFT-GAN) and the baseline (CV-GAN). Each video was generated using the same caption for both methods. The generated videos have a resolution of $64\times64$ pixels and a duration of 32 frames.}
   \label{fig:result}
   %\vspace{-8mm}
\end{figure}

\section{Evaluation}
We evaluated the proposed method qualitatively, via visual inspection of the results, and quantitatively, via an ablation study and a user study. We investigated the effectiveness of our encoding of action-appearance captions using different configurations of the two-stage architecture of CFT-GAN, as well as a single-stage architecture baseline, namely Conditional VGAN (CV-GAN). CV-GAN is a conditional video generation method based on VGAN (Section \ref{sec:related}), in which a caption is encoded in the same way as CFT-GAN (feature extraction and concatenation with latent variables).
Then, we used our self-constructed dataset to train our method and the baseline.

\subsection{Qualitative evaluation}
\label{sec:quali}
Figure \ref{fig:result} shows examples of the generated videos compared to the single-stage baseline. The videos generated by CV-GAN do not reflect the content of the caption properly, that is, the appearance of the person, the action the person is executing, and the background. On the other hand, the proposed method is able to better reflect the contents of the caption. We believe this is because our Conditional TextureGAN uses the caption to generate the background and the foreground separately and, therefore, the appearance of the video can be better controlled. Also, we believe that using Conditional FlowGAN to generate the optical flow from the caption allows us to control better the action we want to reflect in the video. In spite of processing the caption in a similar manner, the baseline is not able to fully use the contents of the action-appearance caption for video generation.
Visually, in the videos generated by CV-GAN, contours tend to be distorted and motion looks less plausible than the videos generated by CFT-GAN. Thus, we can infer the importance of dividing the video generation process into motion and appearance, not only for reflecting the contents of the caption, but also for improving the realism of the generated videos.

\subsection{Quantitative evaluation}
To evaluate objectively to what extent our method reflects the captions content, we used a distance metric between our videos and the original video with the same caption in the dataset. Since the GAN-based generated videos are, by definition, different from the original, we do not expect low distance values; instead we will focus on the difference between each configuration. The root-mean-square distance (RMSD) between two videos is defined as
\begin{equation}
\textrm{RMSD}=\sqrt{\sum\limits_{f=1}^F \sum\limits_{p=1}^P \frac{(p_d-p_g)^2}{P\times F}}
\end{equation}
where $F$ is the number of frames in the video, $P$ is the number of pixels in one frame, $p_d$ is the RGB value of pixel $p$ in the dataset frame, and $p_g$ is the RGB value of pixel $p$ in the generated frame. This measure is also similar to the endpoint error used to compare two optical flows. In our case, this distance increases if there are incoherences in the appearance and motion of the compared videos. Since the dataset and generated videos do not always have the same $F$, when comparing two videos, the longer video is subsampled uniformly to match the lower $F$. Similarly, the video with the higher resolution gets downscaled to the lowest $P$.

\begin{figure}[t]
\begin{center}
  	\includegraphics[width=\hsize]{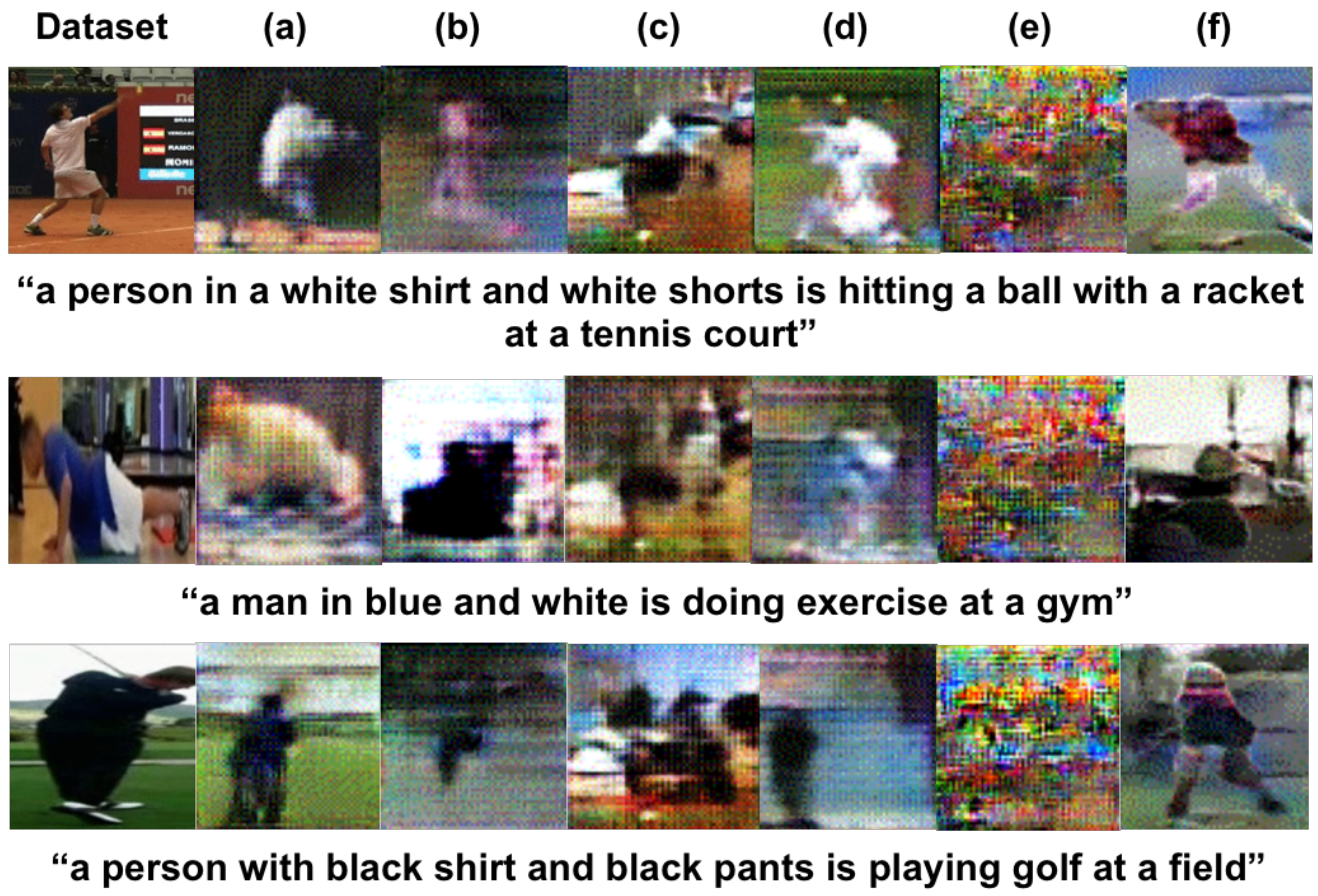}
\end{center}
   \vspace{-1mm}
   \caption{Ablation study. Examples of videos generated with a different way of encoding caption features. Notation of (a-f) is the same as in Table \ref{tb:abla}}
   \label{fig:abla}
   \vspace{-3mm}
\end{figure}

\begin{table}[!h]
\vspace{2mm}
\begin{center}
  \begin{tabular}{|>{\small}l||>{\small}c|} \hline
  Configuration&Distance\\ \hline \hline
  \textbf{(a) Proposed (CFT-GAN)} & \bf{111.03} \\ \hline
  (b) No caption in texture generator & 193.06 \\ \hline
  (c) No caption in tex. gen. foreground only & 170.26 \\ \hline
  (d) No caption in tex. gen. background only & 172.15 \\ \hline
  (e) No caption in flow generator & 221.15 \\ \hline
%  (f) No caption encoding at all & 221.15 \\ \hline
  (f) Single-stage caption encoding (CV-GAN) & 150.68 \\ \hline
  \end{tabular}
  \vspace{1mm}
  \caption{Ablation study of our method. We analyze the contribution of each caption encoding by measuring the distance (RMSD, lower is better) between the generated videos and the original videos in the dataset (averaged).}
  \label{tb:abla}
  \end{center}
  \vspace{-4mm}
\end{table}

Firstly, we performed an ablation study in order to determine the contribution of each caption encoding to the generated video. We randomly selected fifty captions from the dataset and generated the videos using different configurations. We repeated this process five times. Table \ref{tb:abla} summarizes the obtained distance values (averaged for all videos), and their respective frame examples can be found in Figure \ref{fig:abla}. When the encoding of caption features is omitted (i.e., replaced by zeros) in the texture generator (b), although motion is present, the generated videos do not reflect the indicated appearance. Furthermore, omitting the caption separately in the foreground (c) and background (d), leads to inconsistent human appearance and background appearance respectively. On the other hand, when the caption encoding is omitted in the flow generator (e), our architecture is not able to generate a plausible video. We believe this is because the texture generator depends on the output of the flow generator. Finally, when trying our encoding in a single-stage generation method, the videos cannot reflect the caption successfully (see also Section \ref{sec:quali}). This shows the effectiveness of our caption encoding for two-stage video generation.

Lastly, we conducted a user study through Amazon Mechanical Turk \cite{amt} to compare the results between the proposed method and the baseline. As in our qualitative evaluation, we compared our method and the baseline in terms of the capability of the method to reflect the content of the input captions, and the realism of the generated videos. For this, we asked 50 unique workers to visualize 50 pairs of videos (CFT-GAN and CV-GAN) generated using same caption and answer the following two questions: (A) ``Which human video looks more realistic?'', (B) ``Which human video looks more appropriate for the caption?''. In total, we obtained 5000 opinions. %We show some examples of AMT questions in Figure \ref{fig:amt}.
The results of the survey (Table \ref{tb:amt}) show that, for both questions, more participants preferred the videos generated by our method instead of the baseline. This is coherent with the results of our ablation study, since our videos are able to reflect the content of our action-appearance captions. However, for the general user, videos are still not realistic enough, and thus, there is not a huge difference between both methods. In order to improve the realism of our method, the frame resolution could be improved by repeating the generation process taking the low resolution video as an input, similarly to StackGAN~\cite{zhang2016stackgan}. The length of our videos could be also increased, while preserving temporal coherency in motion and appearance. These would require extending our current dataset. We plan to tackle this issues in our future work.

%\begin{figure*}[t]
%\begin{center}
%  	\includegraphics[width=\hsize]{figure/acmmm2018_amt.eps}
%\end{center}
  % \vspace{-2mm}
  % \caption{Some of examples of evaluation test on AMT \cite{amt}.}
 %  \label{fig:amt}
%\end{figure*}

\begin{table}[t]
\vspace{2mm}
\begin{center}
  \begin{tabular}{|c|c|} \hline
  Question&Prefer CFT-GAN over CV-GAN\\ \hline \hline
  A& \bf{58.88\%} \\ 
  B& \bf{55.48\%} \\ \hline
  \end{tabular}
  \end{center}
  \caption{Quantitative evaluation results of our method via AMT \cite{amt}. We show the generated videos and its corresponding captions to the AMT workers, and asked them to select between our method (CFT-GAN) and the baseline (CV-GAN) in two questions. Question A: ``Which human video looks more realistic?'', Question B: ``Which human video looks more appropriate for the caption?''}
  \label{tb:amt}
  \vspace{-3mm}
\end{table}

\section{Conclusions}
We proposed CFT-GAN, a novel automatic video generation method, which encodes action-appearance captions into a two-stage pipeline.
CFT-GAN consists of two GANs; Conditional FlowGAN and Conditional TextureGAN, to reflect a variety of actions (i.e., \textit{doing sit ups}, \textit{playing tennis}) and appearances (i.e., \textit{blue shorts}, \textit{white shirt}...) according to the caption introduced as a condition.
Our experimental results demonstrate that a two-stage structure allows reflecting better the contents of the input action-appearance caption.
To the best of our knowledge, this is the first GAN-based video generation method capable of controlling both the appearance and motion of human action videos using a text caption.
In addition, we constructed a new video dataset for video generation from action-appearance captions, which will serve as source data for future research in this field. Our future work includes improving the resolution and duration of the generated videos to increase their realism, as well as other improvements such as video generation with moving backgrounds.

\section*{Acknowledgement}
We would like to thank Takuhiro Kaneko, Takuya Nanri, Hiroaki Yamane, and Yusuke Kurose for helpful discussions. This work was partially supported by JST CREST Grant Number JPMJCR1403, Japan, and partially supported by the Ministry of Education, Culture, Sports, Science and Technology (MEXT) as "Seminal Issue on Post-K Computer."

{\small
\bibliographystyle{ieee}
\bibliography{egbib}

\begin{thebibliography}{10}\itemsep=-1pt

\bibitem{amt}
{Amazon Mechanical Turk}.
\newblock https://www.mturk.com.

\bibitem{cai2018deep}
H.~Cai, C.~Bai, Y.-W. Tai, and C.-K. Tang.
\newblock Deep video generation, prediction and completion of human action
  sequences.
\newblock In {\em Proc. European Conference on Computer Vision}, pages
  366--382, 2018.

\bibitem{goodfellow2014generative}
I.~Goodfellow, J.~Pouget-Abadie, M.~Mirza, B.~Xu, D.~Warde-Farley, S.~Ozair,
  A.~Courville, and Y.~Bengio.
\newblock Generative adversarial nets.
\newblock In {\em Proc. Conference on Neural Information Processing Systems},
  pages 2672--2680, 2014.

\bibitem{xu2018attngan}
Z.~Hao, X.~Huang, and S.~Belongie.
\newblock {AttnGAN: Fine-grained text to image generation with attentional
  generative adversarial networks}.
\newblock In {\em Proc. Conference on Computer Vision and Pattern Recognition},
  pages 1316--1324, 2018.

\bibitem{hao2018controllable}
Z.~Hao, X.~Huang, and S.~Belongie.
\newblock Controllable video generation with sparse trajectories.
\newblock In {\em Proc. Conference on Computer Vision and Pattern Recognition},
  pages 7854--7863, 2018.

\bibitem{he2018probabilistic}
J.~He, A.~Lehrmann, J.~Marino, G.~Mori, and L.~Sigal.
\newblock Probabilistic video generation using holistic attribute control.
\newblock In {\em Proc. European Conference on Computer Vision}, pages
  466--483, 2018.

\bibitem{ioffe2015batch}
S.~Ioffe and C.~Szegedy.
\newblock Batch normalization: Accelerating deep network training by reducing
  internal covariate shift.
\newblock In {\em Proc. International Conference on Machine Learning}, pages
  448--456, 2015.

\bibitem{isola2016image}
P.~Isola, J.-Y. Zhu, T.~Zhou, and A.~A. Efros.
\newblock Image-to-image translation with conditional adversarial networks.
\newblock In {\em CVPR}, 2017.

\bibitem{kaneko2018generative}
T.~Kaneko, K.~Hiramatsu, and K.~Kashino.
\newblock Generative adversarial image synthesis with decision tree latent
  controller.
\newblock In {\em Proc. Conference on Computer Vision and Pattern Recognition},
  pages 6606--6615, 2018.

\bibitem{kingma2014adam}
D.~Kingma and J.~Ba.
\newblock Adam: A method for stochastic optimization.
\newblock {\em arXiv:1412.6980}, 2014.

\bibitem{klein2015associating}
B.~Klein, G.~Lev, G.~Sadeh, and L.~Wolf.
\newblock Associating neural word embeddings with deep image representations
  using fisher vectors.
\newblock In {\em Proc. Conference on Computer Vision and Pattern Recognition},
  pages 4437--4446, 2015.

\bibitem{marwah2017attentive}
T.~Marwah, G.~Mittal, and V.~N. Balasubramanian.
\newblock Attentive semantic video generation using captions.
\newblock In {\em Proc. International Conference on Computer Vision}, pages
  1435--1443, 2017.

\bibitem{mikolov2013distributed}
T.~Mikolov, I.~Sutskever, K.~Chen, G.~S. Corrado, and J.~Dean.
\newblock Distributed representations of words and phrases and their
  compositionality.
\newblock In {\em Proc. Conference on Neural Information Processing Systems},
  pages 3111--3119, 2013.

\bibitem{odena2017conditional}
A.~Odena, C.~Olah, and J.~Shlens.
\newblock Conditional image synthesis with auxiliary classifier gans.
\newblock In {\em Proc. International Conference on Machine Learning}, pages
  2642--2651, 2017.

\bibitem{ohnishi2018ftgan}
K.~Ohnishi, S.~Yamamoto, Y.~Ushiku, and T.~Harada.
\newblock Hierarchical video generation from orthogonal information: Optical
  flow and texture.
\newblock In {\em Proc. AAAI Conference on Artificial Intelligence}, pages
  1--9, 2018.

\bibitem{pan2017create}
Y.~Pan, Z.~Qiu, T.~Yao, H.~Li, and T.~Mei.
\newblock To create what you tell: Generating videos from captions.
\newblock In {\em Proc. ACM on Multimedia Conference}, pages 1789--1798, 2017.

\bibitem{perronnin2007fisher}
F.~Perronnin and C.~Dance.
\newblock Fisher kernels on visual vocabularies for image categorization.
\newblock In {\em Proc. Conference on computer vision and pattern recognition},
  pages 1--8, 2007.

\bibitem{radford2015unsupervised}
A.~Radford, L.~Metz, and S.~Chintala.
\newblock Unsupervised representation learning with deep convolutional
  generative adversarial networks.
\newblock In {\em Proc. International Conference on Learning Representations},
  pages 1--16, 2016.

\bibitem{reed2016generative}
S.~Reed, Z.~Akata, X.~Yan, L.~Logeswaran, B.~Schiele, and H.~Lee.
\newblock Generative adversarial text to image synthesis.
\newblock In {\em Proc. International Conference on Machine Learning}, pages
  1060--1069, 2016.

\bibitem{revaud:hal-01142656}
J.~Revaud, P.~Weinzaepfel, Z.~Harchaoui, and C.~Schmid.
\newblock {EpicFlow: Edge-preserving interpolation of correspondences for
  optical flow}.
\newblock In {\em Proc. Conference on Computer Vision and Pattern Recognition},
  pages 1164--1172, 2015.

\bibitem{ronneberger2015u}
O.~Ronneberger, P.~Fischer, and T.~Brox.
\newblock U-net: Convolutional networks for biomedical image segmentation.
\newblock In {\em Proc. International Conference on Medical Image Computing and
  Computer-Assisted Intervention}, pages 234--241, 2015.

\bibitem{saito2017temporal}
M.~Saito, E.~Matsumoto, and S.~Saito.
\newblock Temporal generative adversarial nets with singular value clipping.
\newblock In {\em Proc. International Conference on Computer Vision}, pages
  2830--2839, 2017.

\bibitem{tulyakov2018mocogan}
S.~Tulyakov, M.-Y. Liu, X.~Yang, and J.~Kautz.
\newblock Mocogan: Decomposing motion and content for video generation.
\newblock In {\em Proc. Conference on Computer Vision and Pattern Recognition},
  pages 1526--1535, 2018.

\bibitem{vondrick2016generating}
C.~Vondrick, H.~Pirsiavash, and A.~Torralba.
\newblock Generating videos with scene dynamics.
\newblock In {\em Proc. Conference on Neural Information Processing Systems},
  pages 613--621, 2016.

\bibitem{wang2016generative}
X.~Wang and A.~Gupta.
\newblock Generative image modeling using style and structure adversarial
  networks.
\newblock In {\em Proc. European Conference on Computer Vision}, pages
  318--335, 2016.

\bibitem{xu2015empirical}
B.~Xu, N.~Wang, T.~Chen, and M.~Li.
\newblock Empirical evaluation of rectified activations in convolutional
  network.
\newblock {\em arXiv:1505.00853}, 2015.

\bibitem{yang2018pose}
C.~Yang, Z.~Wang, X.~Zhu, C.~Huang, J.~Shi, and D.~Lin.
\newblock Pose guided human video generation.
\newblock In {\em Proc. European Conference on Computer Vision}, pages
  204--219, 2018.

\bibitem{zhang2016stackgan}
H.~Zhang, T.~Xu, H.~Li, S.~Zhang, X.~Huang, X.~Wang, and D.~Metaxas.
\newblock {StackGAN}: Text to photo-realistic image synthesis with stacked
  generative adversarial networks.
\newblock In {\em Proc. International Conference on Computer Vision}, pages
  5908--5916, 2017.

\bibitem{zhang2013actemes}
W.~Zhang, M.~Zhu, and K.~G. Derpanis.
\newblock From actemes to action: A strongly-supervised representation for
  detailed action understanding.
\newblock In {\em Proc. International Conference on Computer Vision}, pages
  2248--2255, 2013.

\bibitem{zhao2018learning}
L.~Zhao, X.~Peng, Y.~Tian, M.~Kapadia, and D.~Metaxas.
\newblock Learning to forecast and refine residual motion for image-to-video
  generation.
\newblock In {\em Proc. European Conference on Computer Vision}, pages
  387--403, 2018.

\end{thebibliography}
}

\end{document}